# Calibrating Generative AI to Produce Realistic Essays for Data Augmentation

Edward W. Wolfe
Justin O. Barber
Pearson

## Abstract

Data augmentation can mitigate limited training data in machine-learning automated scoring engines for constructed-response items. This study seeks to determine how well three approaches to large language model prompting produce essays that preserve the writing quality of the original essays and produce realistic text for augmenting ASE training datasets. We created simulated versions of student essays, and human raters assigned scores to them and rated the realism of the generated text. The results of the study indicate that the (a) "predict next" prompting strategy produces the highest level of agreement between human raters regarding simulated essay scores, (b) "predict next" and "sentence" strategies best preserve the rated quality of the original essay in the simulated essays, and (c) "predict next" and "25 examples" strategies produce the most realistic text as judged by human raters.

## 1 Introduction

Automated scoring engines (ASEs) use machine learning to predict human essay scores. ASEs are a promising supplement to human scoring of essays because they are quicker and reduce scoring costs, facilitating the inclusion of constructed-response items in large-scale assessments. When creating an operationally deployable ASE model, examinee responses to an essay prompt are assigned scores by human raters based on a multi-point scoring rubric, and a machine learning engineer (MLE) uses those responses to develop an ASE model. During that process, the MLE splits the available data (i.e., the response text and the associated scores) into a set used to "train" the engine (a training dataset) and another set used to evaluate the performance of the resulting ASE model (a validation dataset).

For this process to be viable, enough responses are needed at each score point to support the ML algorithms. As such, operational deployment of ASE models can be stymied by the cost and time required to collect human scores and the sparsity of observations at some score points. These limitations may result in small ASE training datasets and can prevent an MLE from identifying an ASE model that meets the score accuracy and reliability requirements set forth by testing stakeholders. Rather than relying on the collection of more human scoring data, an MLE may be able to employ data augmentation (DA) techniques to generate simulated responses that have characteristics that are sufficiently realistic to make them useful for training an ASE.

One drawback of DA techniques is that they may produce unnatural or unrealistic responses. Some DA techniques (usually employing generative AI) tend to increase the sophistication of student responses and to introduce covariate shift, which occurs when the distribution of input features (in this case, the textual responses) changes over time while the relationship between the input features and the target variable (the scores) remains constant (Murphy, 2023). As a result, DA performed with generative AI sometimes leads to a mismatch between the training data and the actual data a model encounters during deployment. Augmenting a training dataset with several AI-generated essays may add samples that distort a model and cause its performance on in-domain samples to either decrease or remain the same. This study examines how to produce DA samples that are as similar to real essays as possible.

## 2 Related Work & Research Questions

### 2.1 Data Augmentation

DA is a regularization strategy that introduces information for the purpose of increasing a machine learning model's capacity to generalize by exposing it to a wider variety of data (Mumuni & Mumuni, 2022). Researchers have considered several DA techniques in the context of ASE training. Word replacement, swapping, and insertion require little effort and can improve modeling outcomes (Chen et al., 2023; Wei & Zou, 2019; Zhang et al., 2020). Other techniques include synonym replacement (Tashu & Horváth, 2022), generative adversarial networks (Park et al., 2022), Large Language Models or LLMs (Gupta, 2023), and backtranslation (Jong et al., 2022).

Other researchers have sought to determine the optimal number of DA samples to produce the best results. Lee et al. (2024) improved performance by adding 15 % more minority-class samples, whereas Cochran et al. (2023) found a 55–89× expansion of small datasets produces optimal results. Feng et al. (2021) achieved peak GPT-2 performance with ~2 synthetic samples per original response, whereas Martin & Graulich (2024) tested ratios from 1:1 augmentation up to fully synthetic training sets. The best results were achieved by combining original student responses with chatbot-generated data.

Barber & Wolfe (2024) augmented statewide writing-assessment essays via TF-IDF word swaps, back-translation, and generative AI. They trained ASEs on datasets with increasing fractions of synthetic responses and compared rater-agreement and insertion density gains to those from extra human scoring, along with projected time and cost. Their results indicated that DA supplementation may result in a modest increase in score point recall when compared to obtaining additional human-assigned scores, that there is little gained by increasing the number of DA responses by more than the number of original responses available, and that the increase is realized at a fraction of the cost and time that would be required to increase sample sizes using human elaboration. They also found that of the three DA methods examined, generative AI modeling was the most effective at improving model performance as measured by human-to-machine agreement.

## 2.2 Research Questions

This paper explores ways to optimize generative AI prompt engineering to supplement ASE training datasets by considering the judged writing quality and perceived realism of simulated essays from the perspective of human raters. The study addresses the following research questions.

1. Which prompt engineering strategies result in simulated essays rated to be most similar to original essays in terms of writing quality?
2. Which prompt engineering strategies result in simulated essays perceived to be most realistic?

# 3 Experiment

## 3.1 Overview

Two types of data were used in this study. First, writing quality scores assigned to real and simulated student essay responses were examined to determine the comparability of measures of interrater agreement and average scores between the essay types. Second, ratings of the realism of simulated essays produced by large language models using different prompt engineering methods were examined to determine comparability of simulated essays relative to real essays.

## 3.2 Process

Data were collected in four phases: essay prompt administration, scoring of real essays by professional raters, creation of simulated essays, and scoring and realism rating of simulated essays by expert raters.

**Essay Prompt Administration.** A single writing assessment prompt was administered to ninth grade students in all public schools in a single midwestern U.S. state as part of an annual summative assessment. The essay prompt elicited informative-explanatory writing from students, and the resulting essays had a median word count of 266 words. The sample of students (N = 6,145) were randomly selected from the full state population of ninth grade students (about 40,000). The state has predominantly rural and suburban communities with a strong majority of White students (about 90%) for whom the highest level of education is high school (over 90%), with an approximately equal split of males and females.

**Scoring of Real Essays by Professionally Trained Raters.** As part of the operational scoring for the statewide summative assessment, each student-written essay was scored according to a four-point scoring rubric that focuses on idea development and organization. Scores were assigned by randomly selecting one professionally trained rater for each essay from a pool of eight raters who were selected from a population of professional raters that is composed of college-educated adults that is representative of the population of college-educated adults in the U.S. with respect to age, sex, and race. All raters were trained to apply the scoring rubric through an extensive training process that included reviewing the scoring rubric, reviewing annotated examples of essays at each score point, practicing applying the scoring rubric to example essays and receiving feedback on the accuracy of the assigned scores, and passing a certification test to confirm that they are able to score with 60% accuracy (i.e., matches a consensus score assigned to each essay by a panel of expert raters).

**Creation of Simulated Essays**. Using the scores assigned by the professionally trained raters, 24 essays at each of the four score points were randomly selected for a total of 96 essays. Each essay was then replicated to produce three simulated versions using a different prompt engineering strategy for each simulated version (288 simulated essays): 25 Examples, Predict Next, and Sentence. In every case, whenever a prompt strategy incorporates real student essays, those essays were selected from essays assigned the same score point as the target essay.

*25 Examples.* This prompt engineering method directs the language model to adopt the persona of an expert linguist. The model is given 24 real student essays and 25 simulated (GPT-generated) essays. The real student essays, which are randomly sorted, serve as positive examples for the language model to emulate. In contrast, the simulated essays are presented as the work of amateur linguists who have failed to create realistic essays and serve as undesirable examples.

The simulated and real student essay pairs are interleaved, with each real student essay paired with a corresponding simulated essay, forming 24 pairs. The 25th pair consists of a final simulated example paired with a place holder for a newly generated essay. The goal is for the language model to produce a newly generated essay that closely resembles the real student essays and demonstrates an improved ability to mimic realistic writing compared to the examples initially created by GPT.

*Predict Next.* This prompt engineering method involves providing the language model with two real student essays. The first essay serves as a point of reference, while the second essay becomes the target that the generated essay should aim to reproduce. The language model is asked to create detailed notes based on the first essay that allow it to predict the content and style of the second essay. Using these notes, the model generates a student essay prompt and corresponding instructions. In a new thread, the language model is presented with the prompt, notes, and guidelines, and it aims to generate an essay that closely resembles the second real student essay. The prompt specifies parameters such as grammatical, spelling, punctuation, and capitalization errors, as well as the number of words, complex sentences, and idiosyncrasies to include, ensuring that the generated essay aligns closely with the characteristics of real student essays.

*Sentence.* This prompt engineering method prompts the language model by iteratively providing it with one sentence at a time. Every sentence comes from a real student essay and is modified substantially in terms of its structure and vocabulary while also replicating the number and kind of grammatical, spelling, punctuation, and capitalization errors in the original sentence.

**Scoring & Realism Rating by Expert Raters**. For each of the 384 essays [96 real + (3 DA methods × 96 simulated essays per method)], one expert rater randomly selected from a pool of four blindly scored the essay using the same four-point scoring rubric used by the professionally trained raters. Expert raters had characteristics similar to the professionally trained raters (i.e., college-educated), but they were selected to serve as supervisors based on past performance on similar scoring projects. Each essay was simultaneously assigned a "real" or "simulated" rating to evaluate the realism of the essay.

### 3.3 Analyses

The analyses focus on the perceived similarity of simulated and real essays in terms of writing quality and perceived realism of the simulated essays.

**Writing Quality (Research Question 1)**. To evaluate the writing quality of simulated essays relative to real essays, level of agreement was compared between the two independent sets of ratings via the proportion of exact agreements and quadratically weighted kappa (QWK). The averages of the scores assigned to each type of essay by the professional and expert raters were also compared.

**Realism (Research Question 2).** To evaluate the realism of simulated essays relative to real essays, the labels (real versus simulated) assigned by expert raters to each essay were compared to that essay's true origin, and we examined the proportion of agreement between ratings and truth.

## 3.4 Results

**Writing Quality (Research Question 1).** To address the first research question (writing quality), the agreement between a pair of writing quality scores (i.e., those assigned by professionally trained raters and those assigned by the expert raters who supervised them during an operational scoring) and the comparability of the average scores assigned to each simulated essay type relative to the score assigned to the original (real) essay were examined conditional on essay version (1 real essay and 3 simulants).

*Score Agreement.* As a point of reference, the overall proportion of agreement between professional and expert raters across all 384 essays equals 0.64, with proportions varying across score points from a low of 0.55 (score point 2) to a high of 0.79 (score point 4). That difference is not statistically significant based on a logistic regression [$\chi^2_{(1)}$ = 3.60, p = 0.06]. QWK across all essays equals 0.69, which would be considered minimally acceptable for operational deployment of an automated scoring model (Williamson et al., 2012).

Figure 1 displays the variation in the proportion of agreement between professional and expert raters across score points separately for each essay type (real versus each type of simulant). The black line (real essay ratings) serves as a reference for interpreting the remaining lines and proximity to that line indicates that a type of simulant better preserved the complexity of the rating decision relative to the original (real) essays. Agreement rates for real essays ranged from 0.54 (score point 3) to 0.92 (score point 4), with an average of 0.71. The Predict Next prompting method produced the highest overall agreement between expert and professional raters (0.72), while the Sentence and 25 Examples methods produced lower overall agreement (0.59 and 0.54, respectively). At the lower score points, professional versus expert agreement on the 25 Examples simulants very closely matched the values observed for real essays, but the agreement rate dropped significantly at score point 4 (to 0.29). The other two prompt engineering methods produced comparable agreement at score points 1 and 4 to that observed with real essays, but considerably different levels of agreement at the two middle score points. This two-way interaction between engineering method and original score point is statistically significant based on logistic regression [$\chi^2_{(3)}$ = 17.90, p = 0.0005].

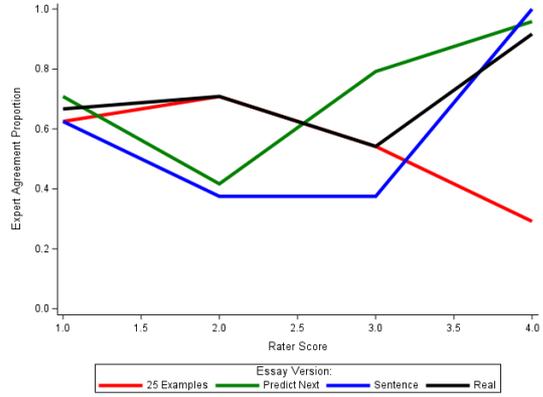

Figure 1: Expert versus Rater Score Agreement by Essay Version.

| Essay Version | QWK |
|---|---|
| Real | 0.75 |
| Predict Next | 0.74 |
| Sentence | 0.68 |
| 25 Examples | 0.58 |

Table 1: QWK by Essay Version.

Table 1 shows the values of QWK by essay version. Compared to the QWK for the real essays (0.75), the Predict Next method produced a comparable level of corrected agreement (0.74), taking into account chance agreement and penalizing bigger rater disagreements. The other two methods produced lower values of QWK. A Mantel-Haenszel test indicates that the variation is not statistically significant [$\chi^2_{(3)}$ = 7.52, p = 0.06]. Regardless, the level of corrected agreement attained for real and Predict Next simulants exceeds the industry standard of 0.70, and the value for the Sentence simulant was close to that threshold (Williamson et al., 2012).

*Average Scores.* Figure 2 displays the variation in the average score assigned by expert raters across a range of score points (assigned by professional raters) separately for each essay type (real versus each type of simulant). The black line (real essay ratings) serves as a point of reference and proximity to that line indicates whether a type of simulant better preserved the writing quality of original (real) essays. The Sentence method produced essays with average scores that were nearly identical to those assigned to the original essays across the scoring range. The Predict Next method produced average scores that were comparable at score points 1, 2 and 4, but the disparity increased at score point 3. Average scores for the 25 Examples method were farthest away from those assigned to the original essays. An analysis of variance indicates that the two-way interaction between the original score and engineering approach is statistically significant although the effect size is small [$F_{(3,376)} = 7.00$, $p = .0001$, $\eta^2 = 0.01$].

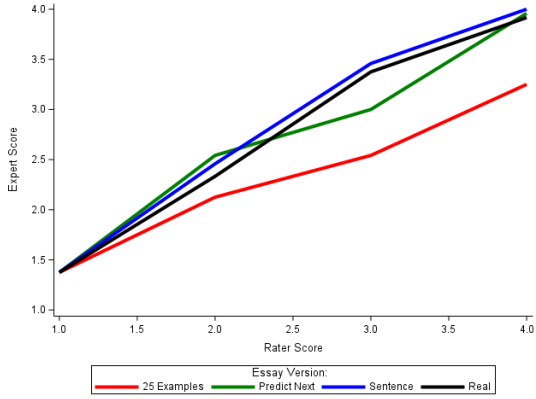

Figure 2: Expert versus Rater Average Scores by Essay Version.

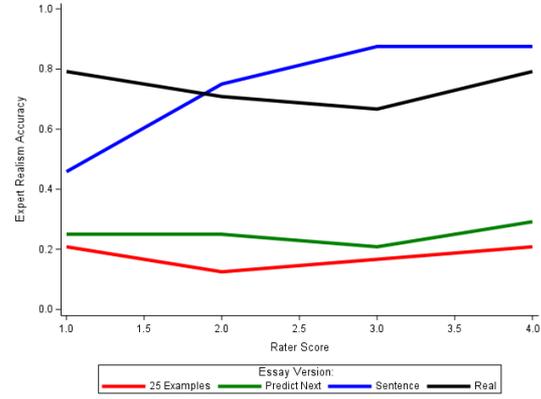

Figure 3: Expert Rater Realism Accuracy by Essay Version.

**Realism (Research Question 2).** To address the second research question (realism), the agreement between the "real" versus "simulated" ratings of expert raters and the true version of the essay by both score point and prompt engineering method were examined. For reference, across all essay types, the proportion of accurate judgments made by expert raters was 0.48, and the accuracy across score points ranged from a low of 0.43 (score point 1) to a high of 0.54 (score point 4). That variation is not statistically significant [$\chi^2_{(1)} = 2.55$, $p = 0.11$]. The value of coefficient kappa for the judgment versus the true status two-way table equals 0.08, indicating that expert rater accuracy, overall, was only 8% better than would be expected by chance.

Figure 3 displays the variation of the proportion of accurate judgments at each original essay score by essay version. Again, the black line (real essay ratings) serves as a point of reference. Values higher than that line indicate that a particular prompt engineering method produced essays for which it was easier to guess the true status relative to real essays, while values lower than that line indicate that the method produced essays that were more difficult to guess their true status. The experts correctly identified real essays at an overall rate of 0.74, ranging from 0.67 (score point 3) to 0.79 (score points 1 and 4). Regarding the transparency of simulated essays, the Sentence method produced the highest level of accuracy in real versus simulated judgments, with an average proportion of 0.74, ranging from 0.46 (score point 1) to 0.88 (score points 3 and 4), which is comparable to real essay rates.

On the other hand, the Predict Next and 25 Examples prompting methods produced essays that were much more difficult to judge accurately, with overall proportions of accurate judgments equaling 0.25 and 0.18, respectively. In both cases, there was very little variation across score points. A logistic regression indicates that this two-way interaction between true status and prompt engineering method is statistically significant [$\chi^2_{(3)} = 8.05$, $p = 0.04$].

## 4 Discussion

### 4.1 Summary and Interpretation of Results

Research question 1 asked which prompt engineering strategies result in simulated essays that are rated to be most similar to original essays in terms of writing quality. We examined expert–professional agreement on individual and mean essay scores. The Predict-Next prompt achieved the highest agreement, comparable to that for real essays. The Sentence and 25 Examples prompting methods produced lower levels of agreement.

However, there is variability of agreement proportions across score points by prompt engineering methods, and these differences are statistically significant. While all prompt engineering methods produced comparable interrater agreement at the lowest score point, the 25 Examples method produced the highest agreement at score point 2, the Predict Next method produced the highest agreement at score point 3, and the Predict Next and Sentence methods produced the highest agreement at score point 4. When chance agreement is considered and a penalty for larger rater disagreements is imposed (i.e., QWK), real essays and those produced with the Predict Next method produce agreement levels consistent with the industry standard of 0.70, and the Sentence method produced values of QWK close to that threshold.

In any case, the fact that rater agreement varies across prompting methods suggests that the features of simulated essays that raters take into account when making a scoring decision may also vary. A prompting method that produces higher values of QWK (e.g., Predict Next) creates simulated essays that are more straightforward for raters to judge reliably, and those accurate decisions seem to be comparable in difficulty to those that raters make regarding real essays.

Regarding the comparability of average writing quality scores across prompting method, the Sentence method produced average scores that are nearly identical to the scores assigned to their real counterparts. The Predict Next method produced average scores that are similar to the real essay counterparts at all score points except score point 3, where those scores were lower than those assigned to the originals. The 25 Examples method produced lower average scores at all score points except score point 1. The variability of the average scores across score points by the prompting methods is statistically significant.

The fact that average writing quality scores vary across prompting methods suggests that the process of simulating essays may slightly change the intended sophistication of the writing relative to their real essay counterparts. Prompts that yield lower mean scores (e.g., the 25-Example method) tend to generate essays with weaker coherence and organization under our rubric. This is important because the purpose of applying DA to ASE training is to provide additional examples that clearly define a particular score category. By decreasing the writing quality of the simulated responses, we run the risk of producing an ASE that does not adequately reproduce the standards that have been set forth in the scoring rubric.

Research question 2 asked which prompt engineering strategies result in simulated essays that are perceived to be most realistic. This question was addressed with analyses of real versus simulated labels that expert raters assigned to each essay. Overall, expert raters were not very successful at identifying the true status of any of the essays, with an overall accuracy rate of only 48%, and they performed only 8% better than chance (i.e., coefficient kappa). Additionally, the accuracy of the experts varied across score points and prompting methods, and that variation was statistically significant. Specifically, they correctly identified real essays about 74% of the time, and they identified essays produced via the Sentence prompting method as simulants at about the same rate. However, simulants produced via the Predict Next and 25 Examples methods were remarkably difficult for raters to correctly identify, and they did so only 25% and 18% of the time, respectively. That is, 75% of the Predict Next simulants were incorrectly labeled as real essays, and 82% of the 25 Examples simulants were incorrectly labeled as real essays. In both cases, there was little variability of accuracy rates across score points.

High detection accuracy (e.g., Sentence prompts) indicates transparently synthetic essays, whereas low accuracy (Predict-Next, 25-Example) suggests essays closer to real writing and thus better suited for ASE data augmentation.

### 4.2 Implications

DA applied to the development of ASEs may potentially allow developers to save time and money by requiring fewer human scores during the engine training process. Previous research has determined that data augmentation can improve the quality of ML models in ASE applications and that there may be an optimal concentration of simulated responses in ASE training datasets (Barber & Wolfe, 2024; Cochran et al., 2023; Feng et al., 2021; Lee et al., 2024; Martin & Graulich, 2024). Additionally, that research has shown that some DA methods produce better ASE training results and that generative AI may produce the best results (Barber & Wolfe, 2024). This leaves an MLE with uncertainty regarding the optimal approach to engineering prompts to produce the most realistic and useful simulants to add to an ASE training dataset.

This research study sought to determine the degree to which three prompt engineering strategies elicit realistic simulated essays for potential use in DA applied to automated essay scoring. Simulated essays that differ in quality or authenticity from real writing may be poor ASE training data. Our findings are largely reassuring: Predict-Next prompts, for instance, produced inter-rater agreement virtually identical to that for real essays. The other two methods produced lower levels of interrater agreement. Additionally, the Sentence prompting method produced average ratings that were nearly identical to those observed for real essays. Predict Next was slightly lower at one score point (score point 3), and the 25 Examples prompting method produced consistently lower scores than those assigned to real essays. Finally, experts were able to identify real essays and simulants created via the Sentence prompting method most of the time. However, they had a difficult time determining that simulants produced using the Predict Next and 25 Examples prompting methods were indeed simulated.

### 4.3 Limitations

This study serves as a first effort to address the question of whether simulated responses to essay prompts are sufficiently similar to real ones to support applications of DA to ASE training, and we acknowledge the limits of the generalizability of the results. Additional research is needed to determine generalizability to other measures of realism (e.g., language parity), content domains (e.g., other than writing), prompt types (e.g., short answers), score point breadth (e.g., 5 score points), languages (e.g., other than English), testing contexts (e.g., other than large-scale summative assessments), and student populations (e.g., other than the U.S. Midwest). The narrow scope of our research design represents only one limited use case which is based on a sector of the world population that may be better represented in the parameterizations of large language models. Hence, we might hypothesize that our results would not be reproducible with populations and application contexts that are disparate from those that we considered. Similarly, we did not consider the potential for the introduction of demographic subgroup bias via a DA supplementation process. Future studies are needed to test whether some prompts over-represent majority language, thereby biasing ASE scores against minority writers.

## References

Barber, J. O., & Wolfe, E. W. (2024, April). *Use of Data Augmentation in Automated Essay Scoring Training Data*. National Council on Measurement in Education, Philadelphia, PA.

Chen, J., Tam, D., Raffel, C., Bansal, M., & Yang, D. (2023). An Empirical Survey of Data Augmentation for Limited Data Learning in NLP. *Transactions of the Association for Computational Linguistics*, *11*, 191–211. https://doi.org/10.1162/tacl_a_00542

Cochran, K., Cohn, C., Rouet, J. F., & Hastings, P. (2023). Improving Automated Evaluation of Student Text Responses Using GPT-3.5 for Text Data Augmentation. In N. Wang, G. Rebolledo-Mendez, N. Matsuda, O. C. Santos, & V. Dimitrova (Eds.), *Artificial Intelligence in Education* (Vol. 13916, pp. 217–228). Springer Nature Switzerland. https://doi.org/10.1007/978-3-031-36272-9_18

Feng, S., Gangal, V., Wei, J., Chandar, S., Vosoughi, S., Mitamura, T., & Hovy, E. (2021). A Survey of Data Augmentation Approaches for NLP. *Findings of the Association for Computational Linguistics: ACL-IJCNLP 2021*, 968–988. https://doi.org/10.18653/v1/2021.findings-acl.84

Gupta, K. (2023). Data Augmentation for Automated Essay Scoring using Transformer Models. *2023 International Conference on Artificial Intelligence and Smart Communication (AISC)*, 853–857. https://doi.org/10.1109/AISC56616.2023.10085523

Jong, Y.-J., Kim, Y.-J., & Ri, O.-C. (2022). Improving Performance of Automated Essay Scoring by Using Back-Translation Essays and Adjusted Scores. *Mathematical Problems in Engineering*, *2022*, 1–10. https://doi.org/10.1155/2022/6906587

Lee, G.-G. "Boaz," Fang, L., & Zhai, X. (2024, April 13). *Improving Machine Scoring Performance with Unbalanced Training Dataset*. 2024 NCME Annual Meeting, Philadelphia, PA.

Martin, P. P., & Graulich, N. (2024). Navigating the data frontier in science assessment: Advancing data augmentation strategies for machine learning applications with generative artificial intelligence. *Computers and Education: Artificial Intelligence*, *7*, 100265.
https://doi.org/10.1016/j.caeai.2024.100265

Mumuni, A., & Mumuni, F. (2022). Data augmentation: A comprehensive survey of modern approaches. *Array*, *16*, 100258. https://doi.org/10.1016/j.array.2022.100258

Murphy, K. P. (2023). *Probabilistic machine learning: Advanced topics*. The MIT Press.

Park, Y.-H., Choi, Y.-S., Park, C.-Y., & Lee, K.-J. (2022). EssayGAN: Essay Data Augmentation Based on Generative Adversarial Networks for Automated Essay Scoring. *Applied Sciences*, *12*(12), 5803. https://doi.org/10.3390/app12125803

Tashu, T. M., & Horváth, T. (2022). Synonym-Based Essay Generation and Augmentation for Robust Automatic Essay Scoring. In H. Yin, D. Camacho, & P. Tino (Eds.), *Intelligent Data Engineering and Automated Learning – IDEAL 2022* (Vol. 13756, pp. 12–21). Springer International Publishing. https://doi.org/10.1007/978-3-031-21753-1_2

Wei, J., & Zou, K. (2019). EDA: Easy Data Augmentation Techniques for Boosting Performance on Text Classification Tasks. *Proceedings of the 2019 Conference on Empirical Methods in Natural Language Processing and the 9th International Joint Conference on Natural Language Processing (EMNLP-IJCNLP)*, 6381–6387. https://doi.org/10.18653/v1/D19-1670

Williamson, D. M., Xi, X., & Breyer, F. J. (2012). A Framework for Evaluation and Use of Automated Scoring. *Educational Measurement: Issues and Practice*, *31*(1), 2–13. https://doi.org/10.1111/j.1745-3992.2011.00223.x

Zhang, D., Li, T., Zhang, H., & Yin, B. (2020). *On Data Augmentation for Extreme Multi-label Classification* (No. arXiv:2009.10778). arXiv. https://doi.org/10.48550/arXiv.2009.10778